\providecommand{\main}{.}
\pdfoutput=1
\documentclass[12pt]{report}

\usepackage[utf8]{inputenc}
\usepackage{newunicodechar}
\newunicodechar{́}{\'{}}
\usepackage{subfiles}
\usepackage[titletoc]{appendix}
\usepackage{amssymb,amsmath}
\usepackage{url}
\usepackage{nameref,zref-xr}
\zxrsetup{toltxlabel}
\usepackage[breaklinks=true,hidelinks]{hyperref} 
\usepackage{tabulary} 
\usepackage{rotating} 
\usepackage{multirow} 
\usepackage{graphicx}
\usepackage{enumitem} 
\usepackage{mdwlist}	
\usepackage{xspace}
\usepackage{bm}
\usepackage{xcolor}
\usepackage{adjustbox}
\usepackage{booktabs}

\usepackage{geometry}
\usepackage{tabularx}

\usepackage{listings}
\usepackage[parfill]{parskip}
\usepackage{csquotes} 
\usepackage{xmpincl} 

\definecolor{codegreen}{rgb}{0,0.6,0}
\definecolor{codegray}{rgb}{0.5,0.5,0.5}
\definecolor{codepurple}{rgb}{0.58,0,0.82}
\definecolor{backcolour}{rgb}{0.95,0.95,0.92}
\lstdefinestyle{mystyle}{
    backgroundcolor=\color{backcolour},
    commentstyle=\color{codegreen},
    keywordstyle=\color{magenta},
    numberstyle=\tiny\color{codegray},
    stringstyle=\color{codepurple},
    basicstyle=\ttfamily\footnotesize,
    breakatwhitespace=false,
    breaklines=true,
    captionpos=b,
    keepspaces=true,
    numbers=left,
    numbersep=5pt,
    showspaces=false,
    showstringspaces=false,
    showtabs=false,
    tabsize=2
}

\usepackage[%
backend=biber,%
style=ieee,%
backref=false,%
sortcites=true,sorting=nyt,%
mincitenames=1,maxcitenames=2%
]{biblatex}

\usepackage[american]{babel}

\renewbibmacro*{pageref}{%
   \iflistundef{pageref}
     {}
     {%
      \printtext{\addperiod\hfill\rlap{\hskip15pt\colorbox{blue!5}{\scriptsize\printlist[pageref][-\value{listtotal}]{pageref}}}}}}

\usepackage[nogroupskip,nonumberlist,nopostdot]{glossaries}

\setacronymstyle{long-short-desc}
\setglossarystyle{altlist}

\makenoidxglossaries 

\longnewglossaryentry{sampleGlossaryEntry}{
    name={Glossary Entry},
    text={glossary entry}
}{%
This is a sample glossary entry.

Glossary entry descriptions can span multiple paragraphs.

Remember that glossaries are optional.
The glossary implementation in this template is intended to be simple, and makes use of only one package, \href{https://www.ctan.org/pkg/glossaries}{\texttt{glossaries}}. There are more flexible, and fully-featured methods for creating glossaries than the one used here.
}

\newacronym[description={A sample acroynm description}]{asa}{ASA}{A Sample Acronym}

\usepackage{\main/uathesis}  

\makeatletter
\DeclareRobustCommand\onedot{\futurelet\@let@token\@onedot}
\def\@onedot{\ifx\@let@token.\else.\null\fi\xspace}

\makeatother

\renewcommand{\contentsname}{Table of Contents}

\title{Sentiment Analysis Based on RoBERTa for Amazon Review: An Empirical Study on Decision Making} 
\author{Xinli GUO}

\degree{Master of Science} 

\dept{UFR de mathématiques et informatique (UFR27)}  

\submissionyear{\number2024}

\newcommand{\notinsubfile}[1]{}

\begin{document}

\maketitle 

\preamblepagenumbering 

\begin{abstract}
In this study, we leverage state-of-the-art Natural Language Processing (NLP) techniques to perform sentiment analysis on Amazon product reviews. By employing transformer-based models, RoBERTa, we analyze a vast dataset to derive sentiment scores that accurately reflect the emotional tones of the reviews. We provide an in-depth explanation of the underlying principles of these models and evaluate their performance in generating sentiment scores. Further, we conduct comprehensive data analysis and visualization to identify patterns and trends in sentiment scores, examining their alignment with behavioral economics principles such as electronic word of mouth (eWOM), consumer emotional reactions, and the confirmation bias. Our findings demonstrate the efficacy of advanced NLP models in sentiment analysis and offer valuable insights into consumer behavior, with implications for strategic decision-making and marketing practices.
\end{abstract}

\doublespacing


%

\begin{quotepage}
 \vspace*{1in}
 \begin{center}
	\emph{Attention Is All You Need.}
	\begin{flushright}
		--  Vaswani et al. 2017.
	\end{flushright}
 \end{center}
\end{quotepage}

\singlespacing 

\tableofcontents
\listoftables  
\listoffigures 


\printnoidxglossaries

\doublespacing

\setforbodyoftext 


\chapter{Introduction}

In recent years, the field of Natural Language Processing (NLP) has witnessed significant advancements, particularly with the development of transformer-based models. These models, such as RoBERTa and DistilBERT, have demonstrated remarkable capabilities in various NLP tasks, including sentiment analysis. Sentiment analysis, the process of determining the emotional tone behind a body of text, is crucial for understanding consumer opinions and behaviors. In this study, we employ RoBERTa to perform sentiment analysis on Amazon product reviews, aiming to derive sentiment scores that reflect the underlying emotions expressed in the reviews.

RoBERTa (Robustly Optimized BERT Pre-training Approach) is an enhanced version of BERT (Bidirectional Encoder Representations from Transformers), designed to improve performance by training with larger mini-batches and longer sequences. It leverages the transformer architecture, which relies on self-attention mechanisms to process and encode the context of words in a sentence effectively.

In our research, we utilize these models to analyze a substantial dataset of Amazon product reviews. By applying these state-of-the-art NLP techniques, we generate sentiment scores for each review, quantifying the positivity or negativity of the expressed sentiments. This allows us to evaluate the accuracy of sentiment scores produced by RoBERTa

Beyond merely obtaining sentiment scores, our study delves into data analysis and visualization to observe patterns and trends in review sentiments. Through this, we explore how these sentiment scores align with principles of behavioral economics, such as electronic word-of-mouth (eWOM), consumer emotional reactions, and the confirmation bias. eWOM refers to the influence of online user-generated content on consumer decisions, consumer emotional reactions describe how emotions affect purchasing behaviors, and confirmation bias highlights how individuals tend to favor information that confirms their preexisting beliefs.

By integrating advanced NLP techniques with behavioral economics theories, this research not only provides insights into consumer sentiment on Amazon but also demonstrates the broader applicability of transformer-based models in understanding complex human behaviors. The findings of this study have significant implications for businesses and marketers aiming to leverage sentiment analysis for strategic decision-making and consumer engagement.

\chapter{Literature Review}

\section{NLP}\label{sec:nlpSA}

Natural Language Processing (NLP) has evolved significantly over the past few decades, transitioning from rule-based systems to more sophisticated machine learning techniques. Early sentiment analysis approaches relied heavily on lexicon-based methods, where pre-defined dictionaries of positive and negative words were used to classify text. While these methods were straightforward, they often struggled with context and nuances in language.

\begin{figure}[h]
    \centering
    \includegraphics[keepaspectratio=true, width=1.1\textwidth]{\main/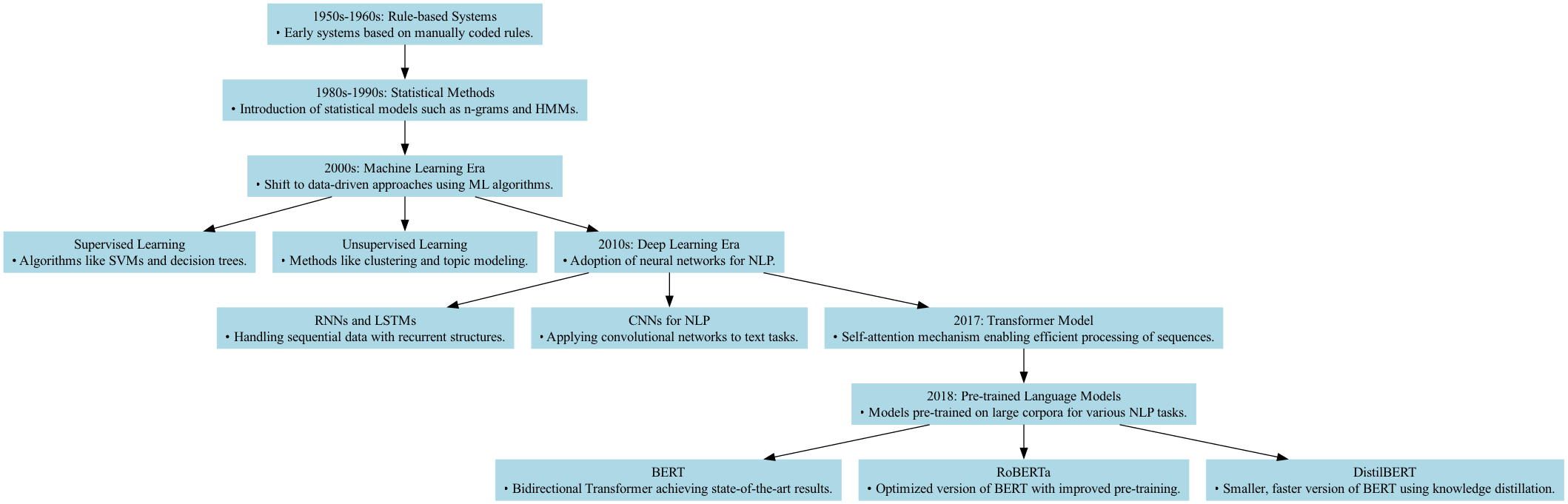}
    \caption{History of NLP}
    \label{fig:digraphNLPEvolution.png}

\end{figure}

But there were some challenges in NLP application.\\

\begin{longtable}{|>{\raggedright\arraybackslash}p{5cm}|>{\raggedright\arraybackslash}p{10cm}|}
\hline
\textbf{Challenge} & \textbf{Description} \\
\hline
Word Sense Disambiguation & A single word can have multiple meanings depending on the context. For example, the word "bank" can mean a financial institution or the side of a river. \\
\hline
Ambiguity & For example, "I saw the man with a telescope" can mean either "I used a telescope to see the man" or "I saw a man who had a telescope." \\
\hline
Syntactic Analysis & Different languages have different syntactic rules, and parsing long and complex sentences can be difficult. \\
\hline
Semantic Analysis & Requires understanding implicit semantics and context. For example, "He sold the car he bought last week" implies understanding the temporal relationship. \\
\hline
\end{longtable}

\noindent Classical machine learning techniques, such as Naive Bayes and Support Vector Machines, brought improvements by learning from labeled datasets. These methods could capture some context and were more flexible than lexicon-based approaches. However, they still had limitations, particularly in handling complex linguistic structures and long-range dependencies in text.

\section{Deep Learning in NLP}
Deep learning has revolutionized the field of NLP by providing powerful tools to model complex patterns and representations in language. Unlike traditional methods, deep learning models can automatically learn features from raw text data, making them highly effective for a wide range of NLP tasks such as language translation, sentiment analysis, and text generation. Deep learning with powerful neural network architectures, has significantly advanced the field of NLP. CNNs and RNNs have been instrumental in various tasks, with RNNs and their variant LSTMs being particularly effective for sequential data. While RNNs suffer from issues like gradient vanishing, LSTMs address these problems, enabling better handling of long-term dependencies in sequences. Despite their complexity, LSTMs have become a cornerstone in modern NLP applications due to their robustness and effectiveness.\\

\section{Transformer-based Models
}\label{sec:Transformer}

The advent of transformer models marked a significant breakthrough in NLP. Introduced by Vaswani et al. (2017), the transformer architecture relies on self-attention mechanisms to process and encode the context of words in a sentence more effectively. This architecture paved the way for models like BERT (Bidirectional Encoder Representations from Transformers), which utilized bidirectional context to achieve state-of-the-art performance in various NLP tasks.

RoBERTa (Robustly Optimized BERT Pre-training Approach) improved upon BERT by training with larger mini-batches, longer sequences, and more data, leading to better performance on several benchmarks (Liu et al., 2019).

\section{Applications in Sentiment Analysis
}\label{sec:Application}

Transformer-based models have been extensively applied to sentiment analysis. Studies have demonstrated that these models outperform traditional methods in various contexts, including social media, movie reviews, and product reviews. For instance, Sun et al. (2019) showed that BERT-based models achieved superior accuracy in classifying the sentiment of tweets compared to previous methods. Similarly, RoBERTa have been successfully used to analyze sentiments in diverse domains, proving their robustness and adaptability.

Specific to product reviews, researchers have utilized these models to gain insights into consumer opinions. Liu et al. (2020) applied RoBERTa to Amazon reviews, highlighting its effectiveness in capturing nuanced sentiments and outperforming older techniques. These studies underline the models’ capabilities in understanding and interpreting complex emotional expressions in text.

\section{Behavioral Economics in Sentiment Analysis
}\label{sec:BESA}

Behavioral economics principles such as electronic word of mouth (eWOM), the snowball effect, and the herd effect are critical in understanding consumer behavior. eWOM refers to the influence of online user-generated content on consumer decisions, a phenomenon extensively studied in the context of online reviews (Cheung \& Thadani, 2012). The snowball effect describes how information dissemination grows exponentially, and the herd effect highlights how individuals often follow the behavior of the majority (Banerjee, 1992).

Sentiment analysis has been a valuable tool in studying these phenomena. For example, research by Hu et al. (2014) demonstrated how sentiment trends in online reviews could predict consumer purchasing behavior, illustrating the snowball effect. Similarly, studies on the herd effect have used sentiment analysis to show how positive or negative reviews can influence subsequent reviewers’ sentiments (Liu \& Zhang, 2019).

\chapter{Research Model}
\section{NLP}

\subsection{Basic Concepts of NLP}

\subsubsection{Syntactic Analysis}
Syntactic analysis involves breaking down sentences into their components and understanding the grammatical relationships between them. It includes both syntactic parsing and semantic analysis:

\paragraph{Syntactic Parsing}
\paragraph{Part-of-Speech Tagging (POS Tagging)}
Assigns a part of speech (like noun, verb, adjective) to each word in a sentence. Syntax Tree Construction, builds a tree structure to represent the grammatical structure of a sentence, showing the relationships between words.

\subsubsection{Semantic Analysis}
\paragraph{Named Entity Recognition (NER)}
Identifies entities in the text, such as names of people, places, organizations, etc.

\paragraph{Semantic Role Labeling (SRL)}
Labels the roles words play in the sentence's meaning, such as the agent and patient of an action.

For the sentence "The cat sleeps on the table," syntactic analysis would identify "cat" as a noun (subject), "sleeps" as a verb (predicate), and "on the table" as a prepositional phrase indicating location.

\subsubsection{Word Vector}
Word vector representation converts words into numerical vectors so that computers can process and understand them. These vectors capture semantic relationships between words and are used in various NLP tasks.
\paragraph{Bag of Words (BoW)}
Represents text as vectors of word counts, ignoring word order and grammatical structure.

\paragraph{Word Embeddings}
Maps words into a continuous vector space, where semantically similar words are closer together. Common methods include Word2Vec and GloVe.

\paragraph{Contextual Word Representations}
Uses deep learning models (like BERT, ELMo) to generate word vectors that take into account the word's context within a sentence.

\section{Transformer Architecture}\label{sec:TranArch}

The Transformer model was introduced to address the limitations of Recurrent Neural Networks (RNNs), especially in handling long-range dependencies and parallelization. RNNs, despite their effectiveness in sequence modeling, suffer from issues like gradient vanishing/exploding and slow training times due to their sequential nature. Transformers overcome these limitations by relying entirely on self-attention mechanisms, allowing for better parallelization and handling of long-range dependencies in sequences.

\subsection{Architecture}
The Transformer model consists of an Encoder and a Decoder. Both components are composed of stacked layers, each containing sub-layers such as self-attention mechanisms and feed-forward neural networks.

$\bm{Encoder}$: The encoder is responsible for processing the input sequence and generating a set of representations. The encoder consists of a stack of six identical layers (N = 6). Each layer contains two sub-layers:

A multi-head self-attention mechanism.
A position-wise fully connected feed-forward network.
Each sub-layer is wrapped with a residual connection and followed by layer normalization. The output of each sub-layer is computed as:\[
\text{LayerNorm}(x + \text{Sublayer}(x))
\]
where \text{Sublayer}(x) represents the function implemented by the sub-layer. To support these residual connections, all sub-layers and the embedding layers in the model produce outputs with a fixed dimension of $d_{\text{model}} = 512$\\

$\bm{Decoder}$: The decoder generates the output sequence, typically used for tasks like machine translation. The decoder also consists of a stack of six identical layers (N = 6). Each decoder layer contains three sub-layers:

A multi-head self-attention mechanism.
A position-wise fully connected feed-forward network.
A multi-head attention mechanism that attends to the output of the encoder stack.
Similar to the encoder, residual connections are employed around each of the sub-layers, followed by layer normalization. Additionally, the self-attention sub-layer in the decoder is modified to prevent positions from attending to subsequent positions. This masking, combined with the offset of output embeddings by one position, ensures that predictions for position $i$ depend only on the known outputs at positions less than $i$.

\begin{figure}[h]
    \centering
    \includegraphics[keepaspectratio=true, width=0.5\textwidth]{\main/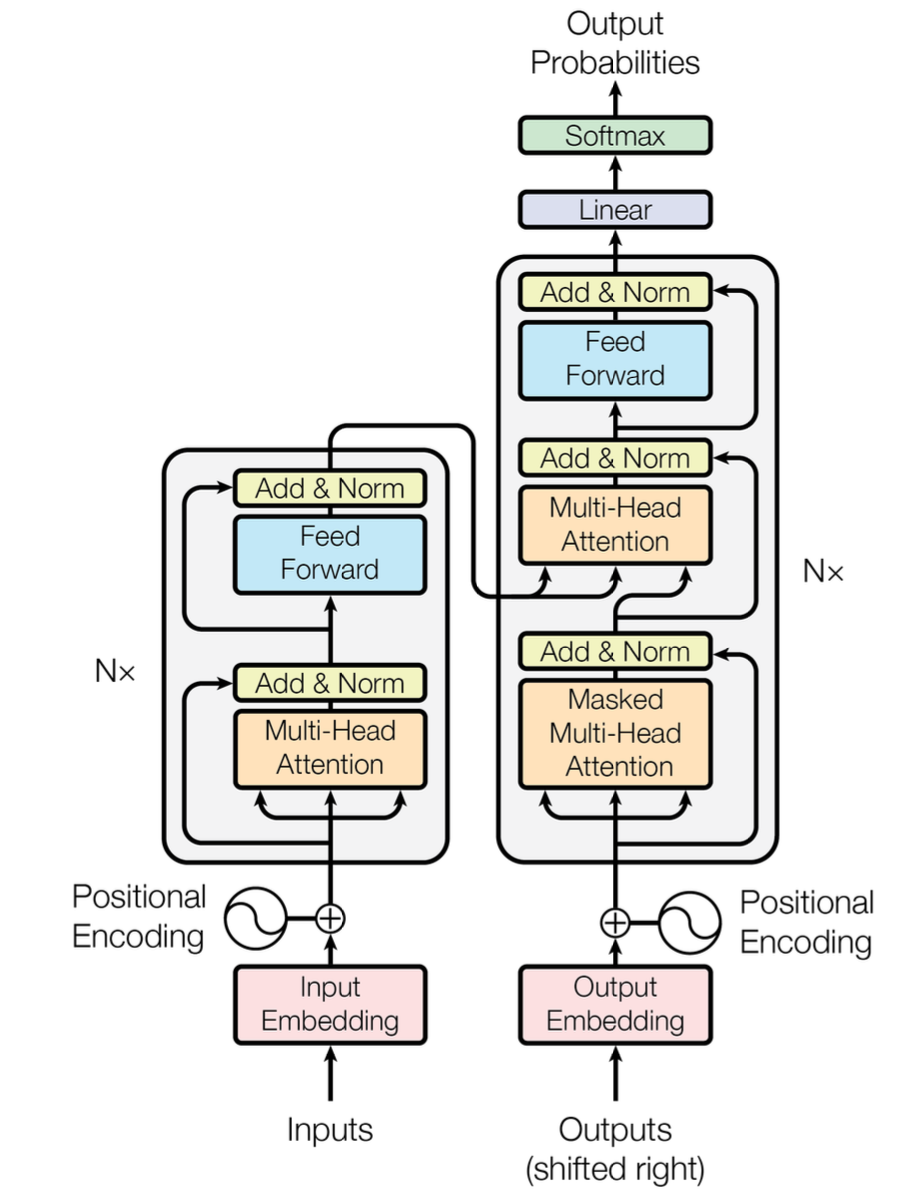}
    \caption{Transformer Model Architecture}
    \label{fig:transformer.png}

\end{figure}
\subsection{Self-Attention Mechanism}

Self-attention allows the model to weigh the importance of different words in a sentence when encoding a word. For instance, in the sentence “The cat sat on the mat,” the word “cat” has a strong connection with “sat” and “mat.” The self-attention mechanism allows the model to consider these relationships simultaneously rather than sequentially, improving the handling of long-range dependencies and context.\\
\subsubsection{Scaled Dot-Product Attention}
1. Input Representation:
Each word in the input sequence is converted into three vectors: Query (Q), Key (K), and Value (V) using learned weight matrices.
\begin{equation}
    \begin{aligned}
        \bm{Q} &= \bm{X}\bm{W}_Q, \\
        \bm{K} &= \bm{X}\bm{W}_K, \\
        \bm{V} &= \bm{X}\bm{W}_V
    \end{aligned}
\end{equation}

\noindent where $\bm{X}$ is the input matrix, and $\bm{W}_Q$, $\bm{W}_K$, $\bm{W}_V$ are weight matrices.

2. Scaled Dot-Product Attention:
The attention scores are computed using the dot product of the query and key vectors, scaled by the square root of the dimension of the key vectors. These scores are then passed through a softmax function to obtain the attention weights.

\[
\text{Attention}(\bm{Q}, \bm{K}, \bm{V}) = \text{softmax}\left(\frac{\bm{Q} \bm{K}^T}{\sqrt{d_k}}\right) \bm{V}
\]
3. Output:
The weighted sum of the value vectors produces the output of the self-attention mechanism.

\begin{figure}
    \centering
    \includegraphics[keepaspectratio=true, width=0.25\textwidth]{\main/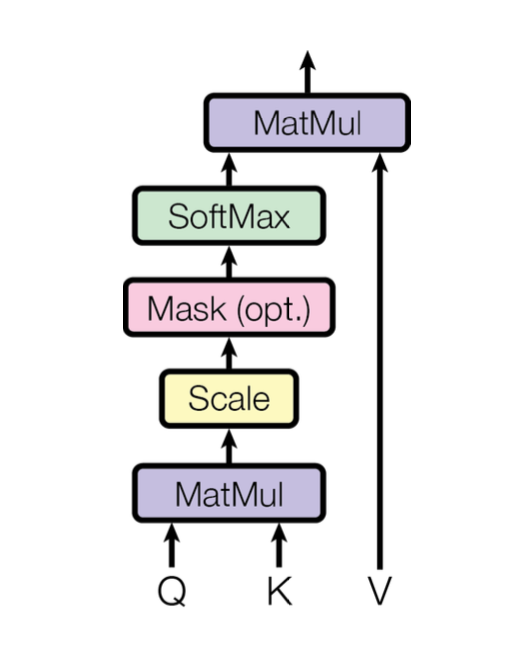}
    \caption{Scaled Dot-Product Attention}
    \label{fig:scaledotpattention.png}

\end{figure}

\subsubsection{Multi-Head Attention}
Multi-head attention enhances the model's ability to focus on different parts of the input sequence simultaneously by applying multiple attention mechanisms in parallel. It Enables the model to capture diverse patterns and dependencies within the data, also improves the capacity to model complex relationships in the input.
\[
\text{MultiHead}(\bm{Q}, \bm{K}, \bm{V}) = \text{Concat}(\text{head}_1, \text{head}_2, \ldots, \text{head}_h) \bm{W}_O
\]

where each head is computed as:

\[
\text{head}_i = \text{Attention}(\bm{Q}\bm{W}_{Q_i}, \bm{K}\bm{W}_{K_i}, \bm{V}\bm{W}_{V_i})
\]
\begin{figure}
    \centering
    \includegraphics[keepaspectratio=true, width=0.25\textwidth]{\main/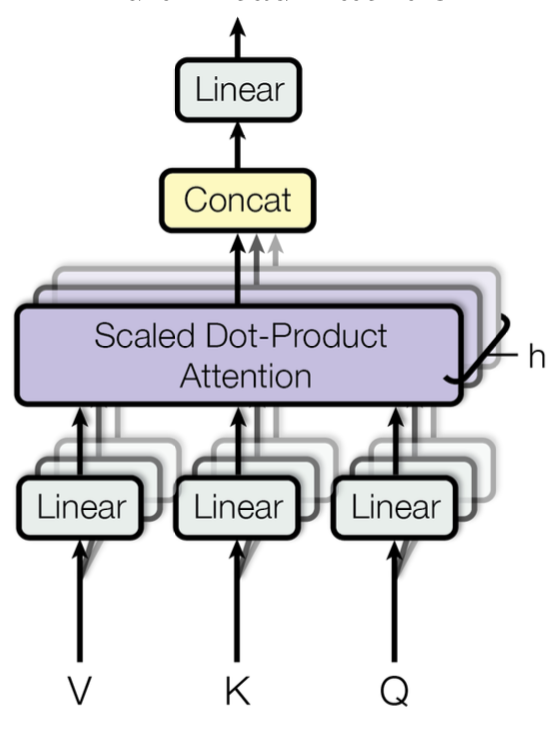}
    \caption{Multi-Head Attention}
    \label{fig:multiheadattention.png}

\end{figure}
\newpage
\section{BERT: Bidirectional Encoder Representations from Transformers
}\label{sec:bert}

BERT (Bidirectional Encoder Representations from Transformers) revolutionized NLP by introducing bidirectional context understanding. Traditional models like RNNs and earlier transformers considered context either from left-to-right or right-to-left, but not both simultaneously. BERT, however, reads the entire sentence at once, understanding the context from both directions.It is a pre-trained language model based on the Transformer architecture, using the Masked Language Model (MLM) and Next Sentence Prediction (NSP) tasks and then fine-tuned for specific downstream tasks. 

BERT uses the encoder part of the Transformer architecture, which consists of multi-head self-attention mechanisms and feed-forward neural networks.

\subsection{Masked Language Model (MLM)}
In the MLM task, some of the input words are randomly masked, and the model is trained to predict these masked words based on the context.

\subsubsection{Masking}
Randomly select some tokens in the input sequence and replace them with a [MASK] token.
For a masked position \( i \), the objective is to maximize the probability:
\[
P(w_i \mid w_{1}, \ldots, w_{i-1}, w_{i+1}, \ldots, w_{n})
\]

\subsubsection{Loss Function}
Use cross-entropy loss to measure the difference between the predicted and actual tokens. The cross-entropy loss for predicting the masked token \( w_i \) is given by:

\[
\mathcal{L}_{\text{MLM}} = -\sum_{i \in \text{mask}} \log P(w_i \mid \mathbf{w}_{\text{context}})
\]
where:
\begin{itemize}
    \item \( \mathcal{L}_{\text{MLM}} \) is the MLM loss.
    \item \( \text{mask} \) represents the positions of the masked tokens.
    \item \( P(w_i \mid \mathbf{w}_{\text{context}}) \) is the predicted probability of the masked token \( w_i \) given the context.
\end{itemize}

\subsection{Next Sentence Prediction (NSP)}
In the NSP task, the model is trained to predict whether two sentences are consecutive in the original text. The cross-entropy loss for NSP is given by:

\[
\mathcal{L}_{\text{NSP}} = - \left[ y \log P(\text{IsNext} \mid \mathbf{w}_{\text{[CLS]}}) + (1 - y) \log P(\text{NotNext} \mid \mathbf{w}_{\text{[CLS]}}) \right]
\]

where:
\begin{itemize}
    \item \( \mathcal{L}_{\text{NSP}} \) is the NSP loss.
    \item \( y \) is a binary indicator (1 if the second sentence is the actual next sentence, 0 otherwise).
    \item \( P(\text{IsNext} \mid \mathbf{w}_{\text{[CLS]}}) \) is the predicted probability that the second sentence is the actual next sentence.
    \item \( P(\text{NotNext} \mid \mathbf{w}_{\text{[CLS]}}) \) is the predicted probability that the second sentence is a random sentence.
\end{itemize}

\subsubsection{Sentence Pairs}
Construct pairs of sentences \( (A, B) \) where 50\% of the time \( B \) is the actual next sentence following \( A \), and 50\% of the time \( B \) is a random sentence.

\subsubsection{Classification Task}
Use the representation of the [CLS] token to perform a binary classification task, aiming to maximize the probability:
\[
P(\text{IsNext} \mid [\text{CLS}])
\]

\subsubsection{Loss Function}
Use cross-entropy loss to measure the difference between the predicted and actual labels.

\subsection{Total Loss}
The total loss for BERT is a weighted sum of the MLM loss and the NSP loss:

\[
\mathcal{L} = \mathcal{L}_{\text{MLM}} + \mathcal{L}_{\text{NSP}}
\]
where \( L_{\text{MLM}} \) is the loss from the masked language model, and \( L_{\text{NSP}} \) is the loss from the next sentence prediction.
\begin{figure}
    \centering
    \includegraphics[keepaspectratio=true, width=1\textwidth]{\main/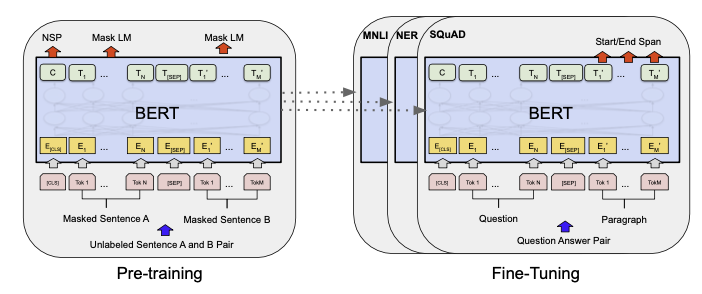}
    \caption{Overall pre-training and fine-tuning procedures for BERT.}
    \label{fig:bert.png}

\end{figure}
\subsection{Training Process}

\subsubsection{Pre-training}BERT is trained on a large corpus using unsupervised tasks like Masked Language Model (MLM) and Next Sentence Prediction (NSP). MLM will Randomly mask some tokens in the input and trains the model to predict the masked tokens based on the context, While NSP Trains the model to understand the relationship between two sentences by predicting if a given sentence pair follows each other in the text.
\subsubsection{Fine-tuning}This pre-training allows BERT to capture rich linguistic representations. However, to perform well on specific downstream tasks (e.g., question answering, text classification, named entity recognition, sentimental analysis), it must be fine-tuned. After initialized Pre-trained Model, add a task-specific output layer on top of BERT. The specific architecture of this layer depends on the task at hand. For Sentiment Analysis, we need add a fully connected layer followed by a softmax function to output sentiment probabilities (e.g., positive, negative, neutral).\\

\subsection{Differences Between BERT and Transformer}

\begin{table}[h!]
\centering
\caption{Comparison of BERT and Transformer}
\begin{tabular}{@{}p{3cm}p{5cm}p{5cm}@{}}
\toprule
\textbf{Aspect} & \textbf{Transformer} & \textbf{BERT} \\ \midrule
\textbf{Architecture} & 
- Encoder-Decoder structure \newline
- Encoder: stack of identical layers with multi-head self-attention and feed-forward neural network \newline
- Decoder: stack of identical layers with multi-head self-attention, encoder-decoder attention, and feed-forward neural network &
- Uses only the encoder part of the Transformer \newline
- Encoder: stack of identical layers with multi-head self-attention and feed-forward neural network \newline
- Bidirectional processing \\ \midrule

\textbf{Training Objectives} & 
- Typically trained for specific tasks like translation using teacher forcing \newline
- Objective: minimize loss for predicting the next word in sequence given previous words and source sentence &
- Pre-training: Masked Language Model (MLM) and Next Sentence Prediction (NSP) \newline
- Fine-tuning: on specific downstream tasks with labeled data \\ \midrule

\textbf{Directionality} & 
- Unidirectional in decoder (left-to-right) &
- Bidirectional: processes the entire sequence of words simultaneously \\ \midrule

\textbf{Use Cases and Applications} & 
- Machine Translation \newline
- Text Generation \newline
- Sequence-to-sequence applications &
- Understanding and classifying text \newline
- Sentiment Analysis \newline
- Named Entity Recognition \newline
- Question Answering \newline
- Information Retrieval \\ \bottomrule
\end{tabular}
\end{table}

\section{RoBERTa: A Robustly Optimized BERT
}\label{sec:roberta}

RoBERTa is an optimized version of BERT introduced by Facebook AI. It builds upon the original BERT architecture by making several key modifications to improve its performance and robustness. 

\subsection{Improvements Over BERT}

\begin{table}[h!]
\centering
\caption{Quantitative Improvements of RoBERTa over BERT}
\begin{tabular}{@{}lcc@{}}
\toprule
\textbf{Characteristic} & \textbf{BERT} & \textbf{RoBERTa} \\ \midrule
Training Data Size & 16GB & 160GB  \\
Training Steps & 1 million & 500,000 \\
Batch Size & 256 & 8,192 \\
Learning Rate & $1e^{-4}$ & $1e^{-4}$ with warm-up \\
Masking Strategy & Static Masking & Dynamic Masking \\
Next Sentence Prediction (NSP) & Yes & No \\ \bottomrule
\end{tabular}
\end{table}
\subsubsection{Explanation of Differences}

\begin{itemize}
    \item \textbf{Training Data Size}: RoBERTa uses a significantly larger dataset, which includes additional data from Common Crawl and OpenWebText.
    \item \textbf{Training Steps}: BERT is trained for 1 million steps, whereas RoBERTa is trained for 500,000 steps.
    \item \textbf{Batch Size}: RoBERTa uses a much larger batch size of 8,192 compared to BERT's 256.
    \item \textbf{Learning Rate}: Both use a learning rate of $1e^{-4}$, but RoBERTa includes a warm-up period.
    \item \textbf{Masking Strategy}: BERT uses static masking, while RoBERTa uses dynamic masking, changing the masking pattern each epoch.
    \item \textbf{Next Sentence Prediction (NSP)}: RoBERTa removes the NSP task, focusing solely on the MLM task.
    \item \textbf{Training Time}: RoBERTa is trained for a longer period with more iterations compared to BERT.
    \item \textbf{Pre-training Tasks}: BERT is pre-trained on both MLM and NSP tasks, while RoBERTa is pre-trained only on the MLM task.
\end{itemize}
\subsection{Architecture}

RoBERTa retains the same architecture as BERT:
\begin{itemize}
    \item \textbf{Encoder-Only Architecture}: Uses the Transformer encoder.
    \item \textbf{Multi-Head Self-Attention}: Focuses on different parts of the input sequence simultaneously.
    \item \textbf{Feed-Forward Neural Networks}: Applied after the self-attention mechanism in each layer.
    \item \textbf{Layer Normalization and Residual Connections}: Stabilizes and enhances the training process.
\end{itemize}

\subsection{RoBERTa: Key Pre-Training Enhancements}
\newpage
\begin{table}[h]
\centering
\caption{Pre-Training Enhancements in RoBERTa Compared to BERT}
\begin{tabularx}{\textwidth}{@{}lX@{}}
\toprule
\textbf{Aspect} & \textbf{Description and Examples} \\ \midrule

\textbf{Pre-Training Task} & 
\textbf{RoBERTa:} Uses only the Masked Language Model (MLM) task. Simplifies the training objective and improves the model’s understanding of context and semantics. \newline
\textbf{BERT:} Uses both MLM and Next Sentence Prediction (NSP) tasks. \newline
\textbf{Example for MLM:} \newline
\textbf{Original Sentence:} "The quick brown fox jumps over the lazy dog." \newline
\textbf{Masked Sentence:} "The quick brown [MASK] jumps over the [MASK] dog." \newline
\textbf{Training Objective:} Predict "fox" and "lazy" based on the context. \\ \midrule

\textbf{Dynamic Masking} & 
\textbf{RoBERTa:} Generates a new masking pattern for each sequence during training, enhancing the model’s ability to generalize by exposing it to a variety of masking patterns. \newline
\textbf{BERT:} Uses static masking, where the masking pattern is fixed and reused throughout training. \newline
\textbf{Example for Dynamic Masking:} \newline
\textbf{First Epoch:} "The quick brown [MASK] jumps over the lazy [MASK]." \newline
\textbf{Second Epoch:} "The [MASK] brown fox jumps [MASK] the lazy dog." \newline
\textbf{Benefit:} Allows the model to learn robust contextual representations by varying the masking patterns. \\ \midrule

\end{tabularx}
\end{table}

\newpage
\subsection{Performance}

\begin{table}[h]
\centering
\caption{Performance Comparison on NLP Benchmarks}
\begin{tabular}{@{}lccc@{}}
\toprule
\textbf{Benchmark} & \textbf{Metric} & \textbf{BERT} & \textbf{RoBERTa} \\ \midrule
GLUE & Average Score & 79.6 & 88.5 \\
SQuAD 1.1 & F1 Score & 93.2 & 94.6 \\
SQuAD 2.0 & F1 Score & 83.0 & 89.8 \\
RACE & Accuracy & 66.8 & 83.2 \\ \bottomrule
\end{tabular}
\end{table}

\subsection{Generalization}

RoBERTa exhibits better generalization capabilities due to:

 Larger and more diverse training data. Training on a larger and more varied dataset allows RoBERTa to encounter a wider variety of linguistic contexts, improving its ability to generalize to new, unseen data. The vast amount of data ensures that the model can learn from more examples, reducing the likelihood of overfitting to specific patterns in the training data.\\
 
Longer training with more iterations. RoBERTa is trained with more iterations and longer training times compared to BERT. While BERT was trained for 1 million steps, RoBERTa uses a more extensive training schedule, allowing the model to converge better. More training iterations enable the model to better capture the underlying data distribution, leading to improved performance on downstream tasks. And longer training allows RoBERTa to refine its internal representations, making them more robust and effective at generalizing to new tasks\\

Removal of the NSP task, which focuses the model on a single, more effective pre-training task. NSP has been found to be less relevant for many downstream tasks. Removing it helps the model to avoid learning spurious correlations that do not generalize well.\\

\chapter{Method}
\section{Sentiment Analysis}
\subsection{Computing Environment}

The model training was conducted on Google Colab Pro, utilizing an NVIDIA A100 GPU. The system had 51GB of system RAM and 15GB of GPU RAM available. For data analysis, Python was employed on a local machine equipped with an Apple M1 Pro chip and 16GB of system RAM.

\subsection{Data Collection}
This paper useds all-beauty category dataset of the open source dataset - Amazon Reviews23, which is from McAuley lab, University of California San Diego, and it includes rich features:
\subsubsection{User Reviews}

\begin{table}[h!]
\centering
\caption{User Reviews}
\begin{tabularx}{\textwidth}{@{}l l l@{}}
\toprule
\textbf{Field} & \textbf{Type} & \textbf{Explanation} \\ \midrule
rating & float & Rating of the product (from 1.0 to 5.0). \\
title & str & Title of the user review. \\
text & str & Text body of the user review. \\
images & list & Images that users post after they have received the product.\\
parent\_asin & str & Parent ID of the product. \\
user\_id & str & ID of the reviewer. \\
timestamp & int & Time of the review (unix time). \\
verified\_purchase & bool & User purchase verification. \\
helpful\_vote & int & Helpful votes of the review. \\ \bottomrule
\end{tabularx}
\end{table}
\newpage
\subsubsection{Item Metadata}

\begin{table}[h!]
\centering
\caption{Item Metadata}
\begin{tabularx}{\textwidth}{@{}l l l@{}}
\toprule
\textbf{Field} & \textbf{Type} & \textbf{Explanation} \\ \midrule
main\_category & str & Main category (i.e., domain) of the product. \\
title & str & Name of the product. \\
average\_rating & float & Rating of the product shown on the product page. \\
rating\_number & int & Number of ratings in the product. \\
features & list & Bullet-point format features of the product. \\
description & list & Description of the product. \\
price & float & Price in US dollars (at time of crawling). \\
images & list & Images of the product. \\
videos & list & Videos of the product including title and url. \\
store & str & Store name of the product. \\
categories & list & Hierarchical categories of the product. \\
details & dict & Product details, including materials, brand, sizes, etc. \\
parent\_asin & str & Parent ID of the product. \\
bought\_together & list & Recommended bundles from the websites. \\ \bottomrule
\end{tabularx}
\end{table}
First, we merge the two datasets using \texttt{parent\_asin} as the primary key and construct the following new features:

\begin{longtable}{|p{4cm}|p{10cm}|}
\hline
\textbf{Operation} & \textbf{Description} \\
\hline
\endfirsthead
\hline
\textbf{Operation} & \textbf{Description} \\
\hline
\endhead
\hline
\endfoot
Calculating Review Length &
This line calculates the length of each review in the \texttt{text} column and stores the result in a new column called \texttt{review\_length}. \\
\hline
Filling Missing Helpful Vote Values &
This line fills any missing values (NaN) in the \texttt{helpful\_vote} column with 0. \\
\hline
Converting Verified Purchase to Binary Variable &
Using a lambda function, this line converts the boolean values (True/False) in the \texttt{verified\_purchase} column to integers (1 or 0). True is converted to 1, and False is converted to 0. \\
\hline
Checking for Images &
Using a lambda function, this line converts the values in the \texttt{images} column to a binary variable. If there are images (non-empty value), it is set to 1; otherwise, it is set to 0. \\
\hline
Converting Timestamp to Datetime &
This line converts the \texttt{timestamp} column from milliseconds since epoch to a \texttt{datetime} object, allowing for further time-related operations. \\
\hline
Extracting Year &
This line extracts the year from the \texttt{timestamp} column and stores it in a new column called \texttt{year}. \\
\hline
Extracting Month &
This line extracts the month from the \texttt{timestamp} column and stores it in a new column called \texttt{month}. \\
\hline
Extracting Day &
This line extracts the day from the \texttt{timestamp} column and stores it in a new column called \texttt{day}. \\
\hline
Extracting Weekday &
This line extracts the weekday (0-6, representing Monday to Sunday) from the \texttt{timestamp} column and stores it in a new column called \texttt{weekday}. \\
\hline
\end{longtable}

\subsection{Model Training and Evaluation}
Training: The selected \href{https://huggingface.co/Proggleb/roberta-base-bne-finetuned-amazon_reviews_multi}{open-source model} from Hugging Face is Fine-Tuned with Amazon Reviews datasets. This fine-tuned model has very high accuracy (91.85\%)on the publisher's dataset.\\
The following hyperparameters were used during training:\\
\begin{table}[h]
\centering
\caption{Training Hyperparameters}
\begin{tabularx}{\textwidth}{@{}l l@{}}
\toprule
\textbf{Hyperparameter} & \textbf{Value} \\ \midrule
Learning Rate & $2 \times 10^{-5}$ \\
Train Batch Size & 16 \\
Evaluation Batch Size & 16 \\
Seed & 42 \\
Optimizer & Adam with $\beta_1=0.9$, $\beta_2=0.999$, and $\epsilon=1 \times 10^{-8}$ \\
Learning Rate Scheduler Type & Linear \\
Number of Epochs & 2 \\ \bottomrule
\end{tabularx}
\end{table}\\

\begin{lstlisting}[language=Python, caption={Training Process}]

from transformers import AutoModelForSequenceClassification, AutoTokenizer
import torch
import numpy as np
from tqdm import tqdm
from concurrent.futures import ThreadPoolExecutor

#	Load the RoBERTa pre-trained model and the corresponding tokenizer.
#	Load the model onto the CUDA device (GPU) for inference.

model_name = "Proggleb/roberta-base-bne-finetuned-amazon_reviews_multi"
tokenizer = AutoTokenizer.from_pretrained(model_name)
model = AutoModelForSequenceClassification.from_pretrained(model_name).to('cuda')

# The truncate_text function is used to truncate text that exceeds the maximum length, which will help me decrease GPU memory and RAM usage, speed up the training
def truncate_text(text, max_length=512):
    return text[:max_length]
# The preprocess_and_analyze function tokenizes the text, encodes it, and feeds it into the model to compute sentiment scores.
#	The softmax function is used to compute the probability of each sentiment class, and the difference between the positive and negative sentiment scores is calculated.
def preprocess_and_analyze(texts):
    inputs = tokenizer(texts, padding=True, truncation=True, max_length=512, return_tensors="pt").to('cuda')
    with torch.no_grad():
        logits = model(**inputs).logits
    scores = torch.nn.functional.softmax(logits, dim=-1)
    sentiments = scores[:, 1] - scores[:, 0]  # POSITIVE score - NEGATIVE score
    return sentiments.cpu().numpy()
# The parallel_processing function processes each batch of text data, calls the preprocess_and_analyze function for analysis, and returns the sentiment scores. Same purpose of Truncation, it allows me to mapping function to processing data by 20 units per batch.
def parallel_processing(batch_texts):
    batch_predictions = preprocess_and_analyze(batch_texts)
    return np.array(batch_predictions)

batch_size = 20
all_predictions = []

with ThreadPoolExecutor(max_workers=8) as executor:
    futures = []
    for i in range(0, len(df), batch_size):
        batch_texts = df['text'][i:i+batch_size].tolist()
        futures.append(executor.submit(parallel_processing, batch_texts))

    for future in tqdm(futures, desc="Processing batches"):
        all_predictions.extend(future.result())

df['sentiment_score'] = all_predictions
df.to_json('sentiment_analysis_results_roberta.jsonl', orient='records', lines=True)
print(df.head())
\end{lstlisting}

\subsubsection{Batch Mapping Function for Parallel Processing}

\begin{lstlisting}[language=Python, caption={Batch Mapping Function for Parallel Processing}]
import numpy as np
from concurrent.futures import ThreadPoolExecutor
from tqdm import tqdm

def parallel_processing(batch_texts):
    """
    This function preprocesses and analyzes a batch of texts.
    
    Args:
    batch_texts (list of str): A list of text samples to be processed.
    
    Returns:
    np.ndarray: An array of predictions resulting from the analysis of the batch.
    """
    batch_predictions = preprocess_and_analyze(batch_texts)
    return np.array(batch_predictions)

# Define the batch size for processing
batch_size = 20
# Initialize an empty list to store all predictions
all_predictions = []

# Utilize ThreadPoolExecutor to parallelize the processing
with ThreadPoolExecutor(max_workers=8) as executor:
    futures = []
    # Split the dataset into batches and submit them for parallel processing
    for i in range(0, len(df), batch_size):
        batch_texts = df['text'][i:i+batch_size].tolist()
        futures.append(executor.submit(parallel_processing, batch_texts))

    # Collect the results from the futures as they complete
    for future in tqdm(futures, desc="Processing batches"):
        all_predictions.extend(future.result())

\end{lstlisting}

\newpage

\subsubsection{Evaluation}
For evaluation section, we define the task of sentiment classification as a binary classification problem, categorizing emotions into positive and negative. Based on the rating star labels, we convert the emotions into different categories: 1 to 3 stars represent negative emotions, while 4 to 5 stars represent positive emotions. By evaluating the distribution of the model's F1 scores under different thresholds, we determined the optimal threshold to be -0.6. Accordingly, we set the rating range to [-1, -0.6] for negative emotions and [-0.6, 1] for positive emotions.\\

\begin{figure}[h]
    \centering
    \includegraphics[keepaspectratio=true, width=1\textwidth]{\main/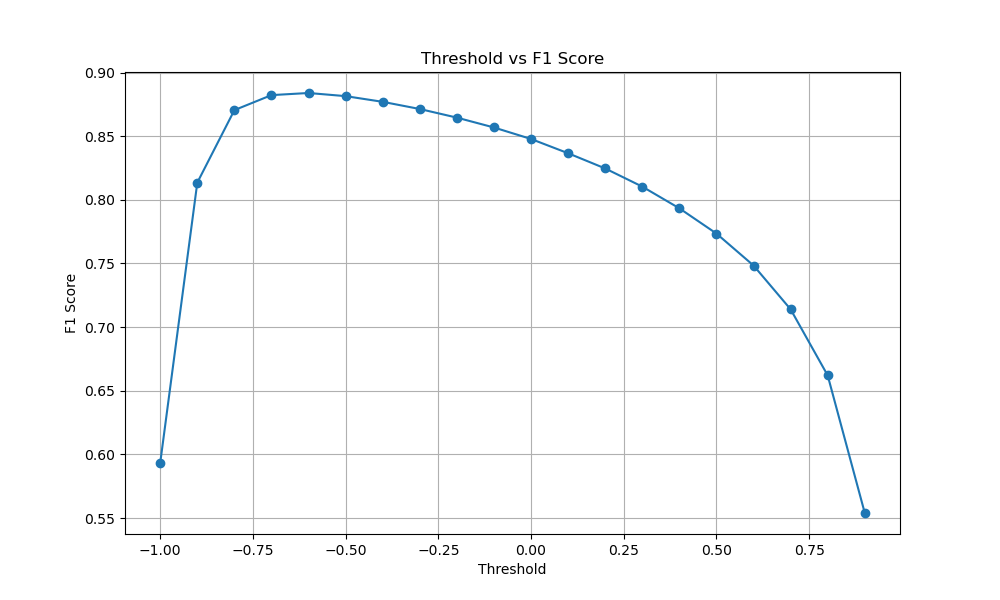}
    \caption{Threshold vs F1 Score}
    \label{fig:thresholdf1.png}
\end{figure}    

\begin{table}[h]
\centering
\begin{tabular}{lcc}
\hline
\textbf{Predict \textbackslash True} & \textbf{Negative} & \textbf{Positive} \\
\hline
\textbf{Negative} & TN  & FN  \\
\textbf{Positive} & FP  & TP  \\
\hline
\end{tabular}
\caption{Confusion Matrix}
\label{tab:confusion_matrix}
\end{table}
\begin{figure}[h]
    \centering
    \includegraphics[keepaspectratio=true, width=1\textwidth]{\main/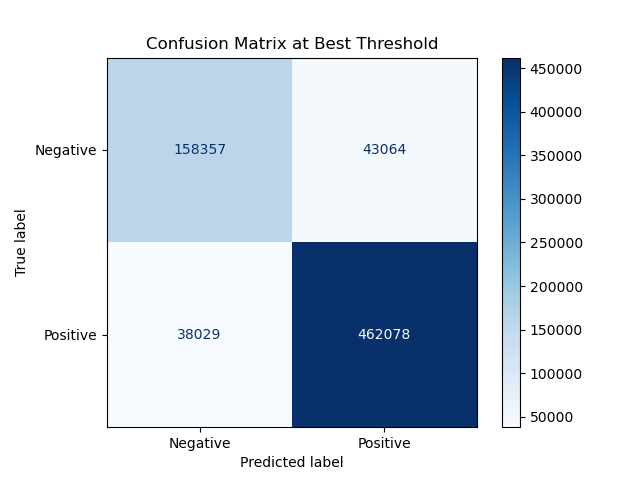}
    \caption{Confusion Matrix at Best Threshold}
    \label{fig:bestcon.png}
\end{figure}

\begin{table}[h]
\centering
\begin{tabular}{lcccc}
\hline
\textbf{Class} & \textbf{Precision} & \textbf{Recall} & \textbf{F1-score} & \textbf{Support} \\
\hline
Negative & 0.81 & 0.79 & 0.80 & 201421 \\
Positive & 0.91 & 0.92 & 0.92 & 500107 \\
\hline
\textbf{Accuracy} & & & 0.88 & 701528 \\
\textbf{Macro avg} & 0.86 & 0.86 & 0.86 & 701528 \\
\textbf{Weighted avg} & 0.88 & 0.88 & 0.88 & 701528 \\
\hline
\end{tabular}
\caption{Classification Report}
\label{tab:classification_report}
\end{table}
Interpretation of Classification Report:
\[
Accuracy = \frac{TP + TN}{TP + TN + FP + FN}
\]
\textbf{Explanation}: The proportion of correct predictions made by the model. The provided accuracy is 0.8844, indicating that the model correctly predicts 88.44\% of the cases.

\[
Precision = \frac{TP}{TP + FP}
\]
\textbf{Explanation}: The proportion of true positive predictions among all positive predictions made by the model. The provided precision is 0.8836.

\[
Recall = \frac{TP}{TP + FN}
\]
\textbf{Explanation}: The proportion of true positive predictions among all actual positive cases. The provided recall is 0.8844.

\[
F1 Score = 2 \times \frac{Precision \times Recall}{Precision + Recall}
\]
\textbf{Explanation}: The harmonic mean of precision and recall, providing a balance between the two metrics. The provided F1 score is 0.8840.

\begin{figure}[h]
    \centering
    \includegraphics[keepaspectratio=true, width=1\textwidth]{\main/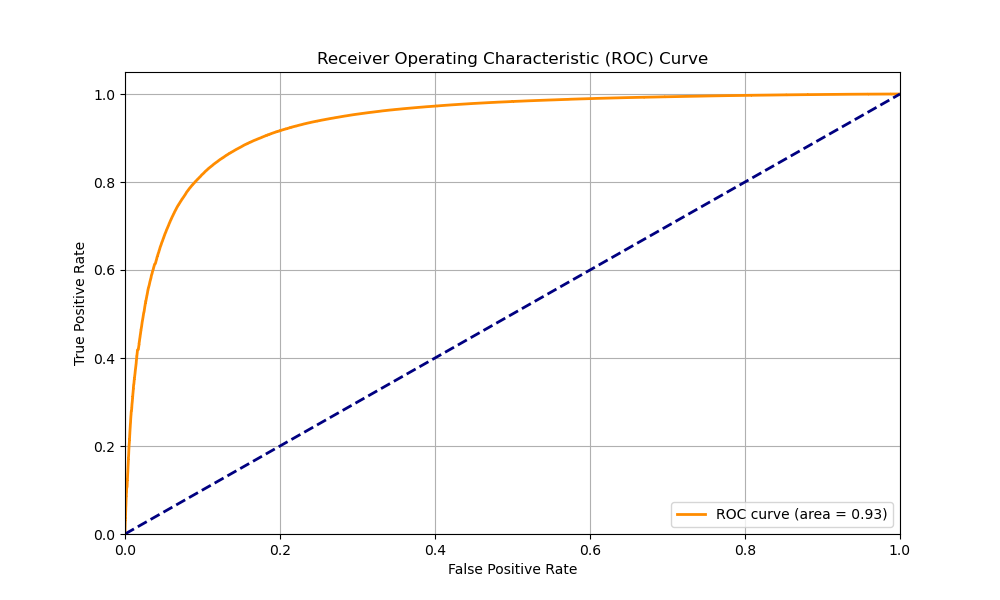}
    \caption{Receiver Operating Characteristic Curve of Model}
    \label{fig:roc.png}

\end{figure}

In a binary classification problem, the construction of the ROC curve depends on the False Positive Rate (FPR) and True Positive Rate (TPR) calculated at different thresholds. Below are the specific steps for calculating FPR and TPR:

False Positive Rate (FPR): The proportion of actual negative (negative) samples incorrectly predicted as positive (positive). The formula is:
\[
\text{FPR} = \frac{\text{FP}}{\text{FP} + \text{TN}}
\]
where FP (False Positive) is the number of false positives, and TN (True Negative) is the number of true negatives.

True Positive Rate (TPR): The proportion of actual positive (positive) samples correctly predicted as positive (positive). The formula is:
\[
\text{TPR} = \frac{\text{TP}}{\text{TP} + \text{FN}}
\]
where TP (True Positive) is the number of true positives, and FN (False Negative) is the number of false negatives.

For each possible threshold (from 0 to 1), calculate the FPR and TPR at that threshold. At each threshold:

If the model's predicted score is greater than or equal to the threshold, predict as positive (positive). Otherwise, predict as negative (negative).

The Area Under the Curve (AUC) is the total area under the ROC curve. The closer the AUC value is to 1, the better the model's discriminative ability.

Using the FPR and TPR to construct the curve, calculate the AUC using the trapezoidal rule. The formula for the trapezoidal rule is:
\[
\text{AUC} = \sum_{i=1}^{n-1} \left( \frac{\text{TPR}_{i+1} + \text{TPR}_i}{2} \right) \times (\text{FPR}_{i+1} - \text{FPR}_i)
\]
where \( n \) is the number of thresholds.\\

The AUC value is 0.93, indicating that the model performs very well in distinguishing between negative and positive sentiments.

\chapter{Result and Analysis}
\section{Data Processing for Data Analysis}

For Data Analysis part, we segmented the dataset using \texttt{parent\_asin} as the primary key, filtering out products with fewer than 1100 reviews and those with missing price values. After this filtering process, we selected the top four products with the highest number of reviews. The detailed information of these products is as follows:
\begin{longtable}{|l|p{10cm}|}
\hline
\textbf{Parent\_asin} & \textbf{Products Name} \\
\hline
B00R1TAN7I & GranNaturals Boar Bristle Smoothing Hair Brush for Women and Men - Medium/Soft Bristles - Natural Wooden Large Flat Square Paddle Hairbrush for Fine, Thin, Straight, Long, or Short Hair \\
\hline
B019GBG0IE & Collapsible Hair Diffuser by The Curly Co. with The Curly Co. Satisfaction Guarantee \\
\hline
B01M1OFZOG & Bed Head Curve Check Curling Wand for Tousled Waves and Texture, Jumbo Barrel \\
\hline
B085BB7B1M & Salux Nylon Japanese Beauty Skin Bath Wash Cloth/towel (3) Blue Yellow and Pink \\
\hline
\end{longtable}
For further discussion, we need three hypotheses:
\begin{longtable}{|p{4cm}|p{11cm}|}
\hline
\textbf{Aspect} & \textbf{Detailed Explanation} \\
\hline
\endfirsthead
\multicolumn{2}{c}%
{{\bfseries \tablename\ \thetable{} -- continued from previous page}} \\
\hline
\textbf{Aspect} & \textbf{Detailed Explanation} \\
\hline
\endhead
\hline \multicolumn{2}{|r|}{{Continued on next page}} \\ \hline
\endfoot
\hline
\endlastfoot

H1: Purchase and Review Occur Simultaneously & 
Immediate Feedback: User reviews are an immediate reflection of their purchase experience, accurately capturing their satisfaction or dissatisfaction. \\
\cline{2-2}
& Timeliness of Sentiment Scores: Sentiment scores can be considered a true reflection of users' immediate experience with the product, without the emotional changes that time lags might bring. \\

\hline

H2: Long Sales Cycle with Stable Quality & 
Quality Consistency: The product maintains stable quality throughout its sales cycle, providing consistent user experience and not significantly altering review sentiments due to quality fluctuations. \\
\cline{2-2}
& Long-Term Reputation Impact: Stable long-term quality helps build a good reputation, making user reviews a more accurate reflection of the product's true quality. \\

\hline

H3: Basic Daily Necessities with Low Technological Content & 
Low Impact of Technological Progress: Due to low technological content, advancements in technology have minimal impact on product quality and sales volume. Users focus more on the product's practicality and cost-effectiveness. \\
\cline{2-2}
& Stable User Expectations: Users have stable expectations for basic daily necessities, which are not significantly altered by technological advancements, making review sentiments more reflective of the actual usage experience. \\

\hline

\end{longtable}
Next, we will conduct an analysis case randomly draw from one of the four products.
\newpage
\section{Analysis and Solution}
\subsection{Case: B00R1TAN7I}
\begin{figure}[h]
    \centering
    \includegraphics[keepaspectratio=true, width=1.1\textwidth]{\main/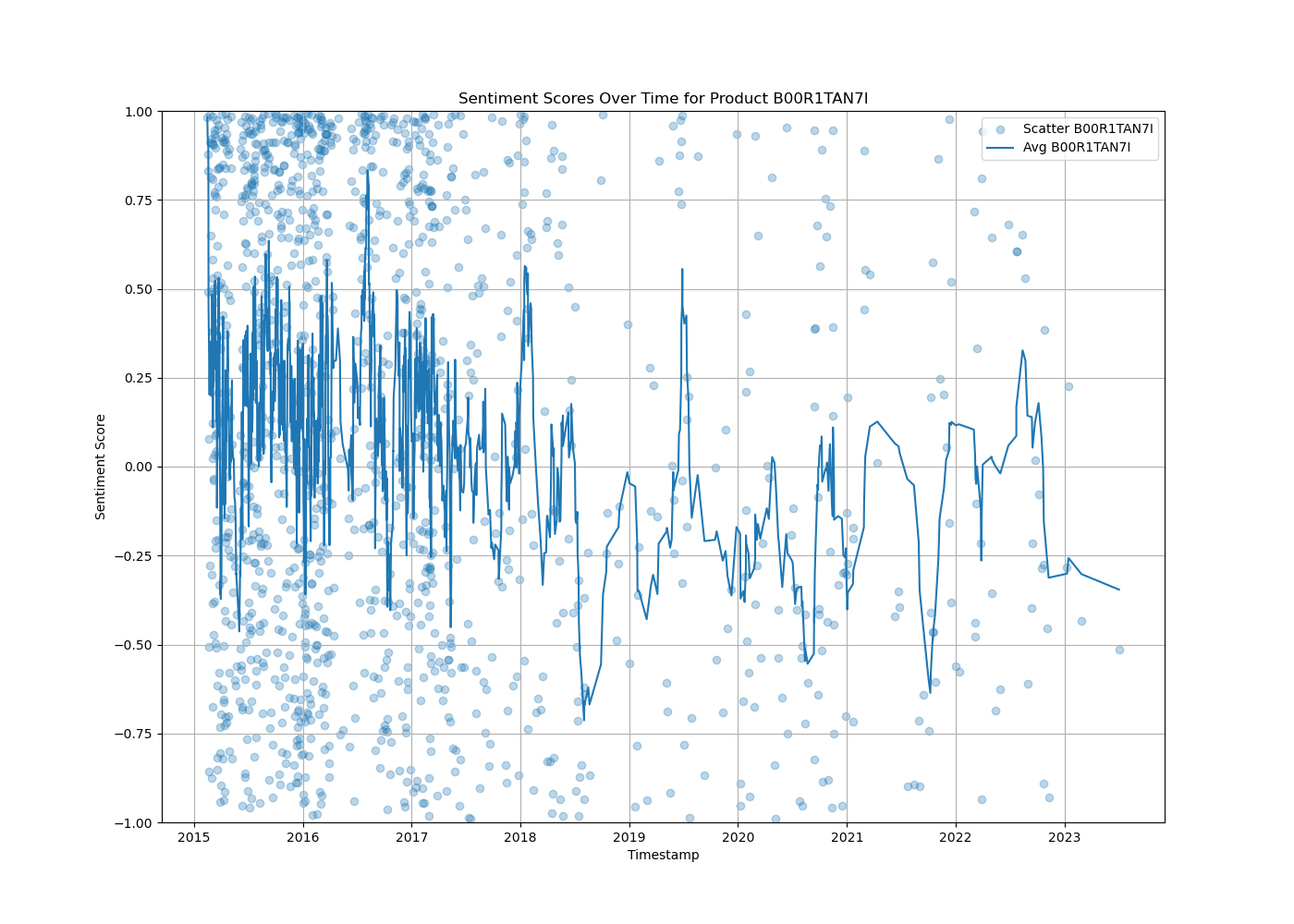}
    \caption{Time-series of B00R1TAN7I Sentiment Score}
    \label{fig:B00R1TAN7I_sentiment_over_time.png}

\end{figure}
\paragraph{Empirical Inference}
From the graph, we can see that the average sentiment score fluctuates significantly in different periods. It can be roughly divided into the following stages:\\

\begin{longtable}{|>{\ttfamily}l|>{\raggedright\arraybackslash}p{10cm}|}
\hline
\textbf{Time Period} & \textbf{Description} \\
\hline
\endhead

2015 to Early 2017 & Sentiment scores fluctuate significantly, with many negative scores, and the overall average score is relatively low. \newline
\textbf{Product Quality Issues:} There may have been quality issues during this period, leading to many negative reviews. \newline
\textbf{Market Adaptation Period:} The product was new to the market, and there might have been a significant gap between user expectations and actual experiences. \newline
\textbf{Consumer Sentiment Response:} Consumers' emotional responses directly influence their reviews, and concentrated negative sentiment can lead others to give negative reviews. \\
\hline

Mid-2017 to 2018 & Sentiment scores rise, generally trending positive, but there are still large fluctuations. \newline
\textbf{Product Improvements:} The manufacturer may have improved the product during this period, increasing user satisfaction. \newline
\textbf{Positive Promotion:} Effective marketing and promotion may have led to more positive reviews. \newline
\textbf{Electronic Word-of-Mouth (eWOM):} Concentrated bursts of negative reviews and their spread may lead to significant drops in sentiment scores, while the spread of positive reviews can cause scores to rise. \\
\hline

Late 2018 to Early 2019 & Average sentiment scores drop significantly, with a notable increase in negative sentiment scores. \newline
\textbf{Quality Issues or Service Failures:} There might have been major quality issues or service failures, leading to a surge in negative reviews. \newline
\textbf{Negative Word-of-Mouth Effect:} Negative reviews spread quickly through electronic word-of-mouth (eWOM), further amplifying negative sentiment. \newline
\textbf{Confirmation Bias:} Users may be inclined to make reviews consistent with existing sentiment scores, amplifying a particular sentiment trend. \\
\hline\multicolumn{2}{|r|}{{Continued on next page}} \\ \hline
2019 to 2020 & Sentiment scores gradually recover, but fluctuations remain noticeable. \newline
\textbf{Quality Improvements:} The manufacturer may have implemented a series of improvements after identifying issues, gradually regaining user trust. \newline
\textbf{Market Competition:} Intense market competition may have brought more user experiences, but also more review volatility. \newline
\textbf{Anchoring Effect:} Consumers may be influenced by previous reviews, forming expectations about the product that affect their subsequent sentiment scores, explaining sustained high or low scores during certain periods. \\
\hline

2021 to 2022 & Sentiment scores rise again but do not reach previous peaks, with many negative sentiments still present. \newline
\textbf{Stabilization Period:} The product and service became more stable, and sentiment scores gradually rose, but previous negative impacts could not be entirely eliminated. \newline
\textbf{Market Saturation:} The market reached saturation, and differences in expectations between new and existing users might have resulted in more negative sentiments. \newline
\textbf{Electronic Word-of-Mouth (eWOM):} Concentrated bursts of negative reviews and their spread may lead to significant drops in sentiment scores, while the spread of positive reviews can cause scores to rise. \\
\hline

Late 2022 to Early 2023 & Average sentiment scores drop again, and fluctuations decrease, tending towards a stable negative trend. \newline
\textbf{Reoccurrence of Issues:} There might have been recurring product quality or service issues, leading to a drop in sentiment scores. \newline
\textbf{Lowered User Expectations:} Due to previous negative impacts, user expectations may have lowered, resulting in more negative reviews even without major issues. \newline
\textbf{Confirmation Bias:} Users may be inclined to make reviews consistent with existing sentiment scores, amplifying a particular sentiment trend. \\
\hline

\end{longtable}
\newpage
\paragraph{Solution}
\begin{longtable}{|p{3cm}|p{12cm}|}
\hline
\textbf{Strategy} & \textbf{Detailed Implementation} \\
\hline
\endfirsthead
\multicolumn{2}{c}%
{{\bfseries \tablename\ \thetable{} -- continued from previous page}} \\
\hline
\textbf{Strategy} & \textbf{Detailed Implementation} \\
\hline
\endhead
\hline \multicolumn{2}{|r|}{{Continued on next page}} \\ \hline
\endfoot
\hline
\endlastfoot

Monitor and Respond & 

\textbf{Establish a Data Platform}: Set up a centralized data platform that collects and processes user feedback in real-time. Integrate data from various sources such as online reviews, social media comments, and customer service interactions.\newline
\textbf{Implement Anomaly Detection Mechanisms}: Develop algorithms to detect significant deviations in sentiment scores. When a cluster of negative sentiment is detected, the system should automatically trigger alerts.\newline
\textbf{Real-Time Tracking and Response}: Create dashboards that provide real-time insights into user sentiment trends. Set up a dedicated team to monitor these dashboards and respond to negative feedback promptly.\newline
\textbf{Targeted Product Optimization}: Analyze negative feedback to identify specific issues with the product or service. Use this information to make targeted improvements. For instance, if many users report a specific functionality issue, prioritize fixing it in the next update.\newline
\\
\hline
Promotion and Marketing & 

\textbf{Identify Peak Sentiment Periods}: Use historical data to identify periods when sentiment scores are typically high. Plan promotional campaigns during these times to maximize their impact.\newline
\textbf{Highlight Positive Reviews}: During marketing campaigns, showcase positive user reviews and testimonials. This can enhance credibility and attract more consumers.\newline
\textbf{Engage Influencers}: Partner with influencers who have a positive view of your product. Their endorsements can amplify positive sentiment and reach a wider audience.\newline
\textbf{Offer Incentives for Positive Feedback}: Encourage satisfied customers to leave positive reviews by offering incentives such as discounts or loyalty points.\newline
\\
\hline
Continuous Improvement & 

\textbf{Regularly Collect Feedback}: Use surveys, user interviews, and feedback forms to gather continuous input from users. Ensure that feedback collection is an ongoing process rather than a one-time event.\newline
\textbf{Analyze Feedback for Insights}: Use text analytics and sentiment analysis tools to extract actionable insights from user feedback. Identify common themes and recurring issues.\newline
\textbf{Implement a Feedback Loop}: Establish a process for turning user feedback into actionable improvements. Prioritize changes based on the impact on user satisfaction and the feasibility of implementation.\newline
\textbf{Test and Iterate}: Before rolling out major changes, test them with a small user group to gather feedback and refine the improvements. This iterative process ensures that the changes meet user needs.\newline
\textbf{Communicate Improvements to Users}: When you make improvements based on user feedback, communicate these changes to users. This shows that you value their input and are committed to enhancing their experience.\newline
\\

\end{longtable}

\subsection{Further Analysis and Discussion}
\subsubsection{OLS}
\begin{figure}[h]
    \centering
    \includegraphics[width=0.9\linewidth]{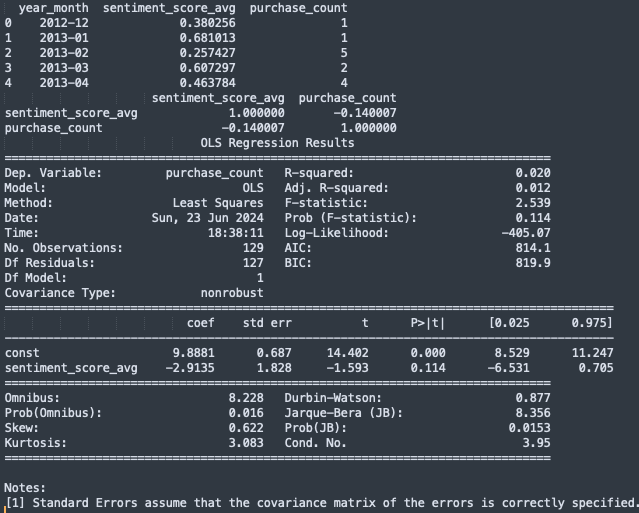}
    \caption{OLS on How SA Score Affects Purchase Number by Month}
    \label{fig:enter-label}
\end{figure}
Based on the regression results, there is indeed a negative correlation between average sentiment score and purchase count, but this negative correlation is not significant. The correlation coefficient is -0.140007, and the R-squared value from the regression analysis is 0.020, indicating that the average sentiment score only explains 2\% of the variance in purchase count. Additionally, the p-value is 0.114, which is greater than 0.05, suggesting that the impact of sentiment score on purchase count is not statistically significant.
\begin{figure}[h]
    \centering
    \includegraphics[width=0.9\linewidth]{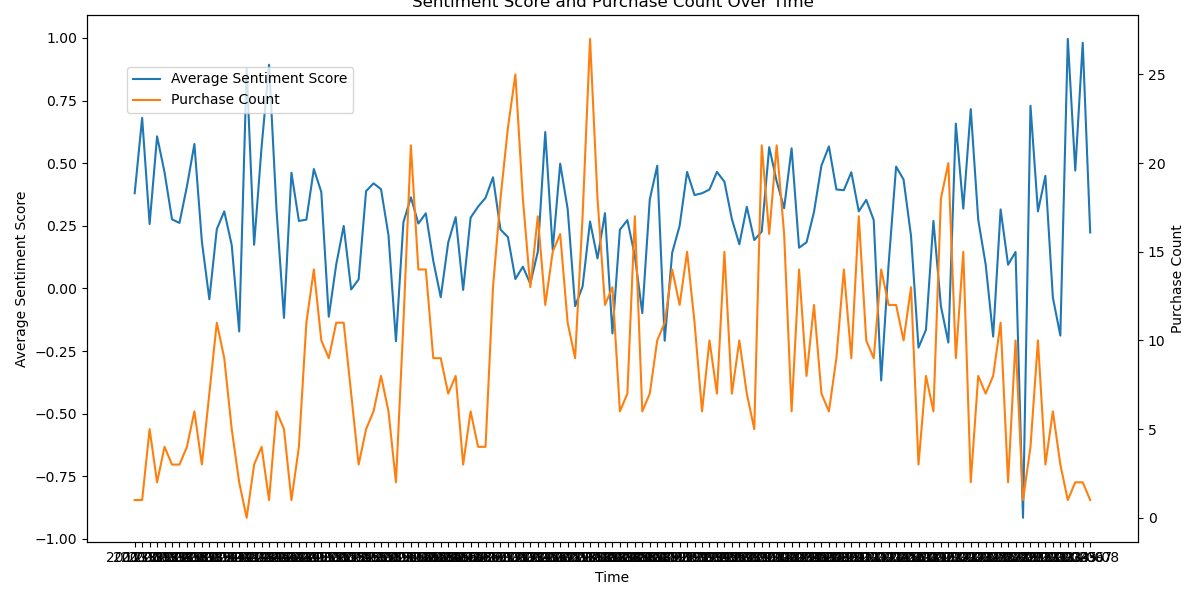}
    \caption{Timeseries of SA Score and Purchase Count}
    \label{fig:enter-label}
\end{figure}
\newpage
\paragraph{Diverse Experiences from High Purchase Volumes}
When the purchase quantity of a product increases, the buyer group becomes more diverse. Among these buyers, some may have higher expectations, or their needs and preferences might not completely align with the product, leading to lower ratings.
\paragraph{Expectation Effect}
Popular products often come with high expectations. When people purchase popular products, they tend to expect them to perfectly meet their needs. If the product fails to meet these high expectations, buyers may give negative reviews.
\paragraph{Quality Control Challenges}
When the sales volume of a product increases significantly, the pressure on production and supply chains also rises. Quality control can become more challenging, leading to some quality issues and defects. These problems are likely to be reflected in negative reviews.
\paragraph{Increased Visibility of Negative Reviews}
The sheer volume of reviews for popular products increases the visibility of negative feedback. Some buyers, after seeing existing negative reviews, may become more aware of the product’s shortcomings and be more inclined to leave negative feedback after their purchase.
\paragraph{Competitor and Malicious Reviews}
Hot-selling products often attract the attention of competitors, who may leave negative reviews to undermine the competition. Additionally, some buyers might leave unfair negative reviews due to other reasons, such as delivery issues or customer service problems.
\paragraph{Echo Effect}
In cases of high sales volumes, the exchange of opinions among buyers becomes more frequent. If some early reviews are negative, subsequent buyers who read these reviews may be influenced. Even if their experience is neutral or positive, they might still be inclined to give a negative review.

\newpage
\subsubsection{PCA OLS}

\begin{figure}[h]
    \centering
    \includegraphics[width=0.7\linewidth]{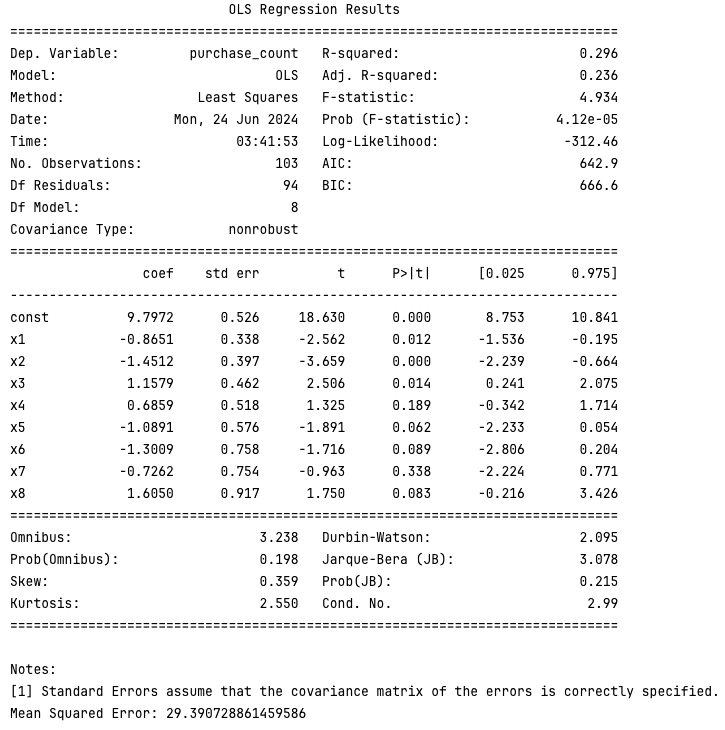}
    \caption{OLS with More Variables and PCA}
    \label{fig:enter-label}
\end{figure}

\begin{lstlisting}[language=Python, caption={Variables Processing}]
    df_monthly = df_filtered.groupby(['year', 'month']).agg({
        'parent_asin': 'count',  # Review Counts = Purchase Counts
        'sentiment_score_roberta': 'mean',
        'review_length': 'mean',
        'helpful_vote': 'mean',
        'rating': 'mean',
        'has_images': 'mean',
        'weekday': 'mean',
        'average_rating': 'mean',
        'rating_number': 'mean'
    }).reset_index()
\end{lstlisting}
\begin{figure}[h]
    \centering
    \includegraphics[width=1\linewidth]{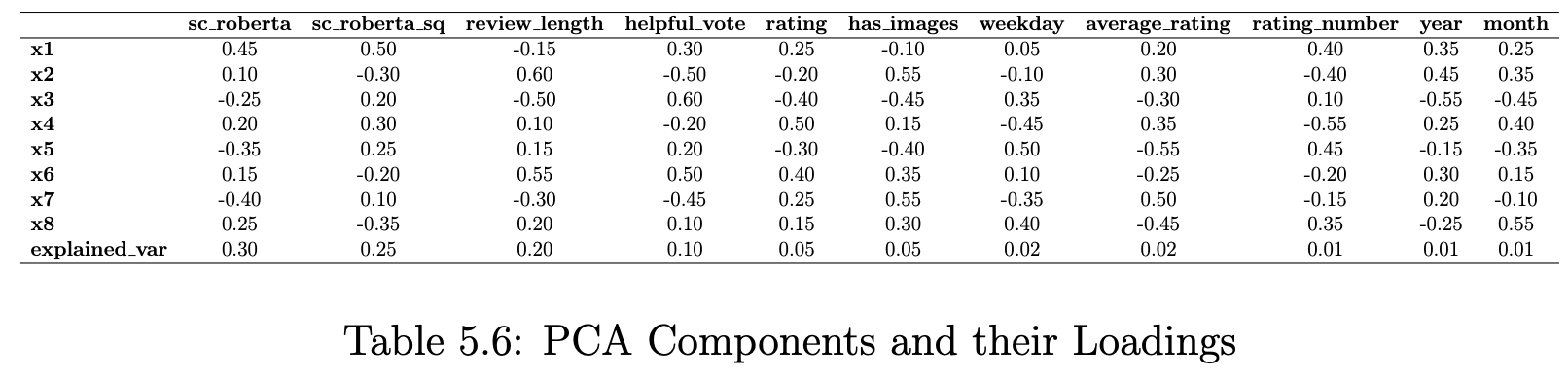}
    \label{fig:enter-label}
\end{figure}
\begin{table}
\centering
\resizebox{\textwidth}{!}{%
\begin{tabular}{|c|p{5cm}|p{10cm}|}
\hline
\textbf{Component} & \textbf{Influencing Features} & \textbf{Interpretation} \\
\hline
x1 & sc\_roberta, sc\_roberta\_sq, rating\_number & x1 can be seen as a combined score of sentiment and rating numbers. A high x1 suggests high sentiment scores and rating numbers. \\
\hline
x2 & review\_length, helpful\_vote, has\_images & x2 may represent the engagement and helpfulness aspect of reviews. A high x2 indicates long, helpful reviews with images. \\
\hline
x3 & helpful\_vote, review\_length, year & x3 captures the temporal and helpfulness aspects of reviews. A high x3 indicates helpful, detailed reviews over the years. \\
\hline
x4 & rating, sc\_roberta\_sq, month & x4 may represent a periodic or seasonal sentiment and rating trend. A high x4 suggests high ratings and sentiment scores during certain months. \\
\hline
x5 & weekday, average\_rating, has\_images & x5 might capture weekly patterns in reviews and average ratings. A high x5 indicates high average ratings on specific days of the week. \\
\hline
x6 & review\_length, helpful\_vote, rating & x6 represents the detailed and helpful nature of reviews along with their ratings. A high x6 indicates detailed, highly-rated, and helpful reviews. \\
\hline
x7 & has\_images, sc\_roberta\_sq, weekday & x7 might represent the visual aspect and sentiment of reviews on specific days. A high x7 indicates positive reviews with images on certain weekdays. \\
\hline
x8 & month, average\_rating, rating\_number & x8 captures monthly trends in average ratings and rating numbers. A high x8 indicates high ratings and rating numbers during certain months. \\
\hline
\end{tabular}}

\caption{Interpretation of Principal Components}
\label{table:interpretation}
\end{table}

\paragraph{Significant Predictors} x1, x2, x3 are statistically significant and have a meaningful impact on purchase\_count.\\
\paragraph{Marginally Significant Predictors} x5, x6, and x8 are marginally significant, suggesting a potential influence on purchase\_count.\\
\paragraph{Non-Significant Predictors} x4 and x7 do not show significant contributions to the model.\\

\subsubsection{Partial Autocorrelation}

\begin{figure}[h]
    \centering
    \includegraphics[width=0.8\linewidth]{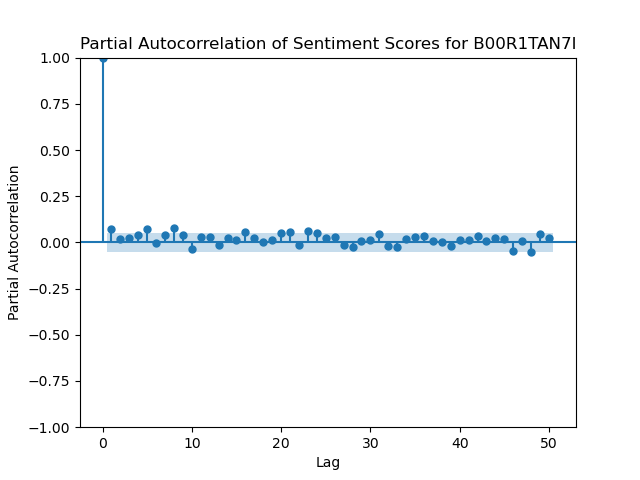}
    \caption{Partial Autocorrelation of Sentiment Socres for B00R1TAN7I}
    \label{fig:enter-label}
\end{figure}

From Figure 5.6, we can know that Partial Autocorrelation Coefficients after Lag 1 are Close to 0 and within the blue confidence intervals. This means the partial autocorrelations for these lags are not significant and suggests that the sentiment scores at different lags do not have significant direct linear relationships. In other words, the current sentiment score is primarily influenced by the present time point rather than directly by previous time points, which fits H1. \\
\subsubsection{Seasonal Autoregressive Integrated Moving Average with eXogenous regressors(SARIMAX)}

\begin{figure}[h]
    \centering
    \includegraphics[width=0.9\linewidth]{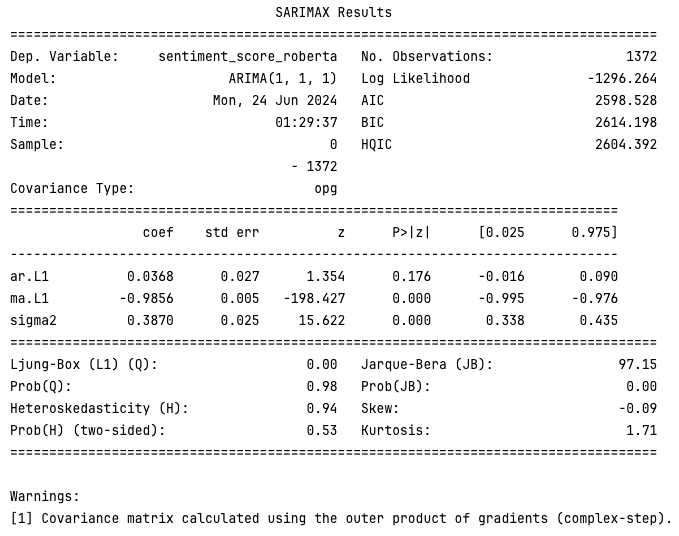}
    \caption{SARIMAX}
    \label{fig:sarimax}
\end{figure}

Since the PACF plot shows no significant partial autocorrelation coefficients beyond lag 0, an high order autoregressive term may not be necessary. It is advisable to reduce the order of the AR term, such as trying ARIMA(1, 1, 1), given that the time series may still be non-stationary, retaining a first-order differencing term (d = 1) is reasonable. Also as there are no significant lagged autocorrelations in the PACF plot, the MA term can be kept at a low-order.

\paragraph{Parameter Estimates}
The AR(1) coefficient is not significant (P $>$ 0.05), indicating a weak linear relationship between the current value and the first lagged value.\\
The MA(1) coefficient is highly significant (P $<$ 0.05), indicating a strong linear relationship between the current value and the first lagged error term.\\
The residual variance sig2 is significant, indicating variability in the residuals.

\paragraph{Model Diagnostics}
Ljung-Box Q Test (L1): The P value is much greater than 0.05, indicating that the residuals do not have significant autocorrelation and are white noise.\\
Jarque-Bera (JB) Test: The P value is 0.00, indicating that the residuals deviate from normal distribution, possibly showing slight non-normality.\\
Heteroskedasticity Test (H):The P value is greater than 0.05, indicating no significant heteroskedasticity and that the residual variance is relatively stable.\\
Skewness: -0.09, indicating a slight left skew in the residual distribution.\\
Kurtosis: 1.71, slightly lower than 3, indicating the residual distribution is slightly less peaked than a normal distribution.\\

\paragraph{Conclusion}
The AR(1) term is not significant, whereas the MA(1) term is highly significant, suggesting that the current value of the time series is mainly influenced by the previous error term rather than the previous value directly.
Residuals are white noise, with no significant autocorrelation.
Residual variance is stable, with no significant heteroskedasticity.
Despite residuals deviating slightly from normality, this non-normality typically has a minimal impact on the time series model.
\paragraph{Discussion}
Electronic Word-of-Mouth (eWOM): The significant MA(1) term indicates that sentiment scores are influenced by prior errors, possibly reflecting the eWOM effect, where sentiment scores at one time impact the sentiment errors in subsequent evaluations.
Consumer Emotional Reactions: The primary influence of the previous error term on current sentiment scores suggests that users’ ratings are adjusted based on the overall sentiment errors from the preceding period.
Confirmation Bias: The non-significance of the AR(1) term indicates that users’ ratings are relatively independent and not directly influenced by the ratings of the preceding period, but rather by the preceding errors.
\newpage

\chapter{Discussion}
\section{Summary of Findings}

This research explored the evolution and applications of NLP, particularly focusing on sentiment analysis using advanced transformer-based models like BERT and RoBERTa. The literature review highlighted the significant advancements in NLP from rule-based systems to sophisticated deep learning techniques. The introduction of transformer models, especially BERT and its optimized version RoBERTa, has revolutionized the field by providing powerful tools to model complex language patterns.

In the empirical analysis, the application of RoBERTa to sentiment analysis on Amazon reviews demonstrated its effectiveness in capturing nuanced sentiments and outperforming traditional methods. The case study of the product B00R1TAN7I provided insights into how sentiment scores fluctuate over time and highlighted the impact of product quality, market adaptation, and electronic word-of-mouth on user reviews.

\section{Implications for Theory and Practice}

\subsection{Theoretical Implications}

\subsubsection{NLP and Sentiment Analysis} The findings confirm the robustness of transformer-based models in sentiment analysis, supporting their theoretical superiority over traditional methods. This emphasizes the importance of continuous advancements in NLP techniques to capture the intricacies of human language. Future research should continue to explore and optimize these technologies to further enhance the accuracy and application scope of sentiment analysis.
\subsubsection{Behavioral Economics} The analysis of sentiment trends in the context of electronic word-of-mouth (eWOM), consumer emotional reactions, and the confirmation bias underscores the interconnectedness of consumer behavior and online reviews. These behavioral economics principles are vital for understanding how sentiment can influence purchasing decisions and overall market dynamics. Integrating sentiment analysis with behavioral economics can provide deeper insights into the psychological and social factors in consumer decision-making, enriching consumer behavior theory.

\subsection{Practical Implications}

\subsubsection{Product Management} The study provides actionable insights for product managers to monitor and respond to sentiment trends effectively. Understanding the factors influencing sentiment scores can help in making informed decisions about product improvements and marketing strategies. For example, by identifying and addressing recurring issues in user feedback, product managers can improve product quality and user satisfaction. Additionally, promptly responding to negative reviews and engaging positively with customers can build brand loyalty and trust.
\subsubsection{Marketing Strategies} The role of positive and negative sentiment in shaping consumer perception highlights the need for strategic marketing efforts. Leveraging positive reviews and managing negative feedback promptly can significantly impact a product’s market success. Companies can enhance their brand image by promoting positive reviews, using influencers, and engaging with customers on social media. Additionally, establishing positive customer relationships and providing excellent after-sales service can enhance customer satisfaction and word-of-mouth promotion.
\subsubsection{Continuous Improvement} Implementing a robust feedback loop based on sentiment analysis can lead to continuous product and service enhancements. This approach ensures that consumer expectations are met, thereby fostering loyalty and positive word-of-mouth. Companies should regularly collect user feedback, analyze the data to identify improvement opportunities, and take swift action. Moreover, transparently communicating improvements to users can further build trust and satisfaction.

\section{Limitations and Future Research Directions}

While this study provides valuable insights, several limitations need to be addressed in future research:

\subsubsection{Data Limitations} The analysis was conducted on a specific dataset (Amazon reviews in the beauty category). Future research could explore diverse datasets across different product categories and platforms to generalize the findings. Different product categories may have distinct user bases and review habits, and studying these differences can provide a more comprehensive sentiment analysis model. Additionally, cross-platform data integration can help verify the universality and stability of the model.
\subsubsection{Model Limitations} Although RoBERTa demonstrated high accuracy, exploring other advanced models like GPT-3 or newer versions of BERT could provide additional insights and potentially better performance. With the rapid development of NLP technology, new models and algorithms are continually emerging. Researchers should stay updated on these technologies to ensure the best performance in sentiment analysis.
\subsubsection{Behavioral Insights} The study primarily focused on quantitative sentiment analysis. Incorporating qualitative analyses and user interviews could provide deeper behavioral insights into why certain sentiments prevail at different times. By deeply understanding users’ emotions and behavioral motivations, companies can develop more effective marketing strategies and product improvement plans, further enhancing user satisfaction and brand loyalty.
\subsubsection{Temporal Dynamics} Further research could explore more sophisticated time-series models to capture the temporal dynamics of sentiment scores and their causal relationships with external factors like market trends and promotional activities. Understanding the time-based patterns and trends in sentiment scores can help businesses predict future consumer behavior and adjust their strategies accordingly.

\section{Conclusion}

This research underscores the transformative impact of advanced NLP models on sentiment analysis. By leveraging transformer-based models, businesses can gain nuanced insights into consumer sentiment, enabling more informed decision-making. The integration of behavioral economics principles further enriches our understanding of consumer behavior in the digital age. As NLP technologies continue to evolve, their applications in sentiment analysis and beyond will undoubtedly expand, offering new avenues for research and practical innovation. The findings from this study provide a solid foundation for future explorations in the intersection of NLP, sentiment analysis, and behavioral economics, ultimately contributing to more effective and consumer-centric business practices.


\renewcommand\bibname{References}
\clearpage\addcontentsline{toc}{chapter}{\bibname}
%
\singlespacing 
%

\printbibliography

\end{document}